\documentclass[sigconf]{acmart}
\acmISBN{}
\acmDOI{}
\acmConference[ICEA2024]{2024 International Conference on Intelligent Computing and its Emerging Applications}{November 28 – 29, 2024}{Tokyo, Japan}
\AtBeginDocument{%
  }

\setcopyright{acmcopyright}
\acmYear{2024}

\acmPrice{15.00}
\acmISBN{978-1-4503-XXXX-X/18/06}
\usepackage{multirow}

\begin{document}

\title{Where Do You Go? Pedestrian Trajectory Prediction using Scene Features}

\author{Mohammad Ali Rezaei$^1$, Fardin Ayar$^1$, Ehsan Javanmardi$^2$, Manabu Tsukada$^2$, Mahdi Javanmardi$^1$}
\email{mohammadalire94@aut.ac.ir,fardin.ayar@aut.ac.ir, ejavanmardi@g.ecc.u-tokyo.ac.jp, mtsukada@g.ecc.u-tokyo.ac.jp, mjavan@aut.ac.ir}
\affiliation{%
\institution{$^1$Department of Computer Engineering, Amirkabir University of Technology, Tehran, Iran \\
$^2$Graduate School of Information Science and Technology, The University of Tokyo, Tokyo, Japan}
\country{}}

\renewcommand{\shortauthors}{Mohammad Ali et al.}

\begin{abstract}
Accurate prediction of pedestrian trajectories is crucial for enhancing the safety of autonomous vehicles and reducing traffic fatalities involving pedestrians. While numerous studies have focused on modeling interactions among pedestrians to forecast their movements, the influence of environmental factors and scene-object placements has been comparatively underexplored. In this paper, we present a novel trajectory prediction model that integrates both pedestrian interactions and environmental context to improve prediction accuracy. Our approach captures spatial and temporal interactions among pedestrians within a sparse graph framework. To account for pedestrian-scene interactions, we employ advanced image enhancement and semantic segmentation techniques to extract detailed scene features. These scene and interaction features are then fused through a cross-attention mechanism, enabling the model to prioritize relevant environmental factors that influence pedestrian movements. Finally, a temporal convolutional network processes the fused features to predict future pedestrian trajectories. Experimental results demonstrate that our method significantly outperforms existing state-of-the-art approaches, achieving ADE and FDE values of 0.252 and 0.372 meters, respectively, underscoring the importance of incorporating both social interactions and environmental context in pedestrian trajectory prediction.

\end{abstract}



\maketitle

\begin{figure}[h]
    \centering
    \includegraphics[width=1\linewidth]{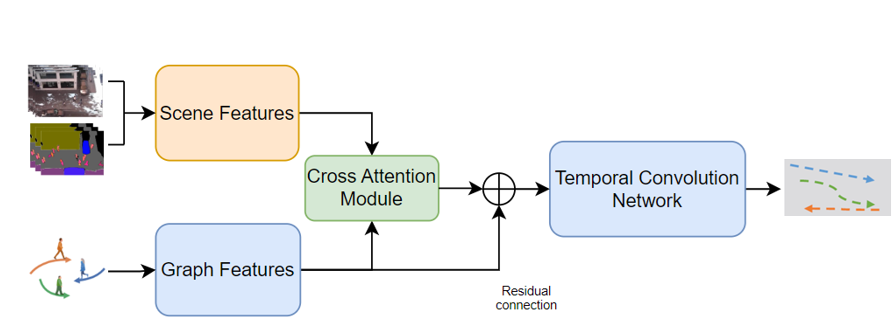}
    \caption{Our model predicts future trajectories by integrating scene and graph features. To achieve this, the approach employs a cross-attention module followed by a Temporal Convolutional Network.}
    \label{fig:figx}
\end{figure}

\section{Introduction}

Advancements in artificial intelligence and machine learning have significantly propelled the field of autonomous vehicles, establishing it as one of the most compelling and challenging areas of research in recent decades \cite{paden2016survey}. Self-driving cars, envisioned as the future of transportation, require sophisticated systems capable of navigating safely and efficiently through complex and dynamic urban environments. A critical challenge for these systems is the accurate prediction of pedestrian trajectories \cite{parekh2022review}.

Predicting pedestrian movement in crowded and dynamic environments is a multifaceted and complex problem that necessitates a detailed analysis of human behavior and interactions with the surrounding environment. Unlike humans, who naturally anticipate others' paths using sensory information and cognitive processing, autonomous vehicles must achieve this capability through mathematical models and sophisticated algorithms. This involves not only understanding individual pedestrian behaviors but also capturing social interactions and adhering to unwritten social norms that people unconsciously follow in communal settings\cite{golchoubian2023pedestrian}.

\subsection{Factors Affecting Pedestrian Trajectory Prediction}

The primary factors influencing pedestrian trajectory prediction include:

\paragraph{Spatial Position of Pedestrians in the Scene}

The spatial position of a pedestrian relative to other pedestrians and elements within the traffic scene is a significant factor \cite{alahi2016social, gupta2018social}. For instance, when two pedestrians approach each other head-on, they instinctively adjust their paths to avoid a collision. Group interactions are also significant; individuals moving in a group establish coordinated movement patterns, influencing each other's trajectories.

\paragraph{Interaction with the Surrounding Environment}

Interactions with the physical environment play a vital role in determining pedestrian movement paths \cite{robicquet2016learning}. Elements such as buildings, street layouts, and physical obstacles constrain movement space and affect decision-making \cite{sadeghian2019sophie}. Understanding these environmental constraints enables prediction models to forecast more logical and feasible paths by excluding areas occupied by obstacles.

\paragraph{Pedestrian Movement Preferences and Intentions}

Information about a pedestrian's speed and acceleration significantly impacts path prediction \cite{mordan2021detecting}. For example, someone moving at a constant speed or running provides cues that are effective for predicting future trajectories. However, fully anticipating a person's intentions is challenging, as sudden factors may cause unexpected changes in their path \cite{rasouli2019pie}, introducing inherent unpredictability.

In this work, to model pedestrian-to-pedestrian interactions and individual trajectory preferences, we employ a Sparse Graph Convolutional Network (SGCN) \cite{mohamed2020social}. To account for interactions between pedestrians and the environment, we introduce a novel feature extraction module that derives scene features from the ETH and UCY datasets \cite{pellegrini2009you}. We utilize an attention mechanism to effectively combine the scene features with the sparse graph features, enhancing the accuracy of pedestrian trajectory prediction.

\section{Related Work}

Humans navigate crowded environments by adhering to unwritten social rules and conventions that govern their interactions with others and the environment\cite{zamboni2022pedestrian}. For example, pedestrians typically avoid passing directly between two individuals engaged in conversation, as doing so would violate social norms. These implicit rules influence pedestrian trajectories and are critical for accurate path prediction models.

Early approaches to modeling pedestrian behavior focused on the Social Force Model \cite{helbing1995social}, which introduced hand-crafted functions based on physical concepts of attraction and repulsion to simulate interactions in traffic scenes. While effective for short-term predictions, the model's computational complexity and limitations in capturing long-term dependencies made it less suitable for applications such as autonomous driving.

The advent of neural networks, particularly Long Short-Term Memory (LSTM) networks \cite{yu2019review}, revolutionized trajectory prediction by effectively modeling temporal dependencies. Alahi et al. \cite{alahi2016social} proposed the Social LSTM, extending traditional LSTMs to account for social interactions by incorporating neighboring pedestrians' hidden states. This approach enabled the modeling of longer sequences and more complex social behaviors.

Subsequent research enhanced these models by integrating collective interactions and environmental contexts. Quan et al. \cite{quan2021holistic} introduced a holistic LSTM framework that encodes collective interactions among pedestrians. Gupta et al. \cite{gupta2018social} developed the Social GAN, leveraging Generative Adversarial Networks to produce socially acceptable trajectories that respect social norms.

Recognizing the importance of environmental factors, recent studies have incorporated scene information into trajectory prediction. Sadeghian et al. \cite{sadeghian2019sophie} proposed SoPhie, an attentive GAN that combines social and physical constraints by integrating scene context through visual features extracted from images. Their work demonstrated that neglecting scene information could lead to decreased prediction accuracy. Similarly, Convolutional Neural Networks (CNNs) have been utilized to capture spatial features of the environment, enhancing the model's understanding of physical constraints \cite{zamboni2022pedestrian}.

Further advancements involve the use of semantic segmentation and attention mechanisms. Syed and Morris \cite{syed2019sseg, syed2023semantic} integrated semantic scene information to refine trajectory predictions, highlighting the benefits of understanding scene semantics.

Graph-based models have also gained prominence for their ability to capture complex interactions among multiple agents. Mohamed et al. \cite{mohamed2020social} introduced Social-STGCNN, utilizing spatio-temporal graph convolutional neural networks to model pedestrian interactions over time. Shi et al. \cite{shi2021sgcn} proposed the Sparse Graph Convolution Network (SGCN), which efficiently models sparse interactions in crowded scenes, leading to improved trajectory prediction performance.

\section{Proposed Method}

We present an approach for accurately predicting pedestrian trajectories by integrating social interactions and environmental context. Our model leverages the Sparse Graph Convolutional Network (SGCN) \cite{shi2021sgcn} to capture pedestrian interactions within spatial and temporal sparse graphs. Additionally, we incorporate a scene feature extraction module and employ a cross-attention mechanism to enhance predictive capabilities.

\subsection{Problem Definition}

Trajectory prediction aims to estimate the future positions of all agents based on their historical states and surrounding scene information \cite{alahi2016social, gupta2018social}. At time $t$, the scene is represented by an image $I_t$. Each pedestrian $i \in [N]$ is characterized by spatial coordinates $(x_i^t, y_i^t) \in \mathbb{R}^2$. The historical positions up to time $t$ are:

\[
X_i^{1:t} = \{ (x_i^\tau, y_i^\tau) \mid \tau = 1, \ldots, t \}, \quad \forall i \in [N],
\]

and the future ground truth positions from time $t+1$ to $T$ are:

\[
Y_i^{t:T} = \{ (x_i^\tau, y_i^\tau) \mid \tau = t+1, \ldots, T \}, \quad \forall i \in [N].
\]

Our objective is to learn a predictive model $f$ parameterized by $W_\Theta$:

\[
\hat{Y}_i^{t:T} = f\left( I_t, X_i^{1:t}, X_{1:N \setminus i}^{1:t}; W_\Theta \right),
\]

where $\hat{Y}_i^{t:T}$ are the predicted future trajectories.

\subsection{Overall Model}

Our model comprises three primary components (Figure~\ref{fig:1}):
\begin{enumerate}
    \item \textbf{Feature Extraction Module}
    \item \textbf{Interaction Module}
    \item \textbf{Cross-Attention Module}
\end{enumerate}

First, the \textit{Feature Extraction Module} processes input frames using Real-ESRGAN \cite{wang2021real} for image enhancement, followed by the OneFormer model \cite{jain2023oneformer} for semantic segmentation to generate semantic maps. Enhanced frames are passed through ResNet-18 \cite{he2016deep} to extract visual features. Semantic maps are processed through additional convolutional layers to obtain semantic features. Visual and semantic features are concatenated and refined via a Multi-Layer Perceptron (MLP) to form a comprehensive scene representation $H_{\text{scene}}$.

Simultaneously, the \textit{Interaction Module} employs the SGCN to capture pedestrian interactions by constructing spatial and temporal sparse graphs from trajectory data. Graph convolutional layers extract features representing motion tendencies and social dynamics.

Next, the \textit{Cross-Attention Module} \cite{lin2022cat} integrates the scene features $H_{\text{scene}}$ with the interaction features $H_{\text{graph}}$. By employing a cross-attention mechanism, the model focuses on relevant environmental factors influencing pedestrian trajectories.
Finally, the fused features are input into a \textit{Temporal Convolutional Network (TCN)} \cite{mohamed2020social} to predict future trajectories.

\subsection{Feature Extraction Module}

To generate high-level environmental representations, we enhance input frames using Real-ESRGAN \cite{wang2021real} to improve image quality. The enhanced images are processed by the pre-trained OneFormer model \cite{jain2023oneformer} for semantic segmentation, producing detailed semantic maps that identify critical scene elements based on the Cityscapes dataset \cite{cordts2016cityscapes}.

We use ResNet-18 \cite{he2016deep} to extract robust visual features from the enhanced frames. Semantic maps are processed through additional convolutional layers to obtain semantic feature vectors. The visual and semantic features are concatenated and refined using an MLP to produce the scene representation $H_{\text{scene}}$.

\subsection{Interaction Module}

The \textit{Interaction Module} utilizes the SGCN \cite{shi2021sgcn} to model pedestrian interactions:

\begin{itemize}
    \item \textbf{Spatial Sparse Graph}: Encodes group interactions at each time step, with nodes representing pedestrians and edges representing significant interactions. The adjacency matrix is sparse and learned during training.
    \item \textbf{Temporal Sparse Graph}: Models motion tendencies over time, connecting temporal instances of each pedestrian. This captures sequential movement patterns.
\end{itemize}

Graph convolutional layers extract features from these graphs, capturing both social dynamics and individual motion patterns. The fusion of spatial and temporal features provides a comprehensive understanding of pedestrian behaviors.

\subsection{Cross-Attention Module}

The \textit{Cross-Attention Module} integrates scene features with interaction features to enhance trajectory prediction. The graph features serve as the \textit{Query} ($Q$), while the scene features provide the \textit{Key} ($K$) and \textit{Value} ($V$):

\begin{equation}
\text{Attention}(Q, K, V) = \text{softmax}\left( \frac{Q K^\top}{\sqrt{d_k}} \right) V,
\label{eq:attention}
\end{equation}

where $d_k$ is the dimensionality of the key vectors. This mechanism allows the model to focus on relevant environmental factors that influence each pedestrian's movement.

The attention output is:

\begin{equation}
H_{\text{attn}} = \text{Attention}(Q, K, V).
\end{equation}

To preserve original interaction information and enhance training stability, a residual connection combines the attention output with the original graph features:

\begin{equation}
H_{\text{fused}} = H_{\text{attn}} + H_{\text{graph}}.
\end{equation}

\begin{figure*}[h]
    \centering
    \includegraphics[width=1\linewidth]{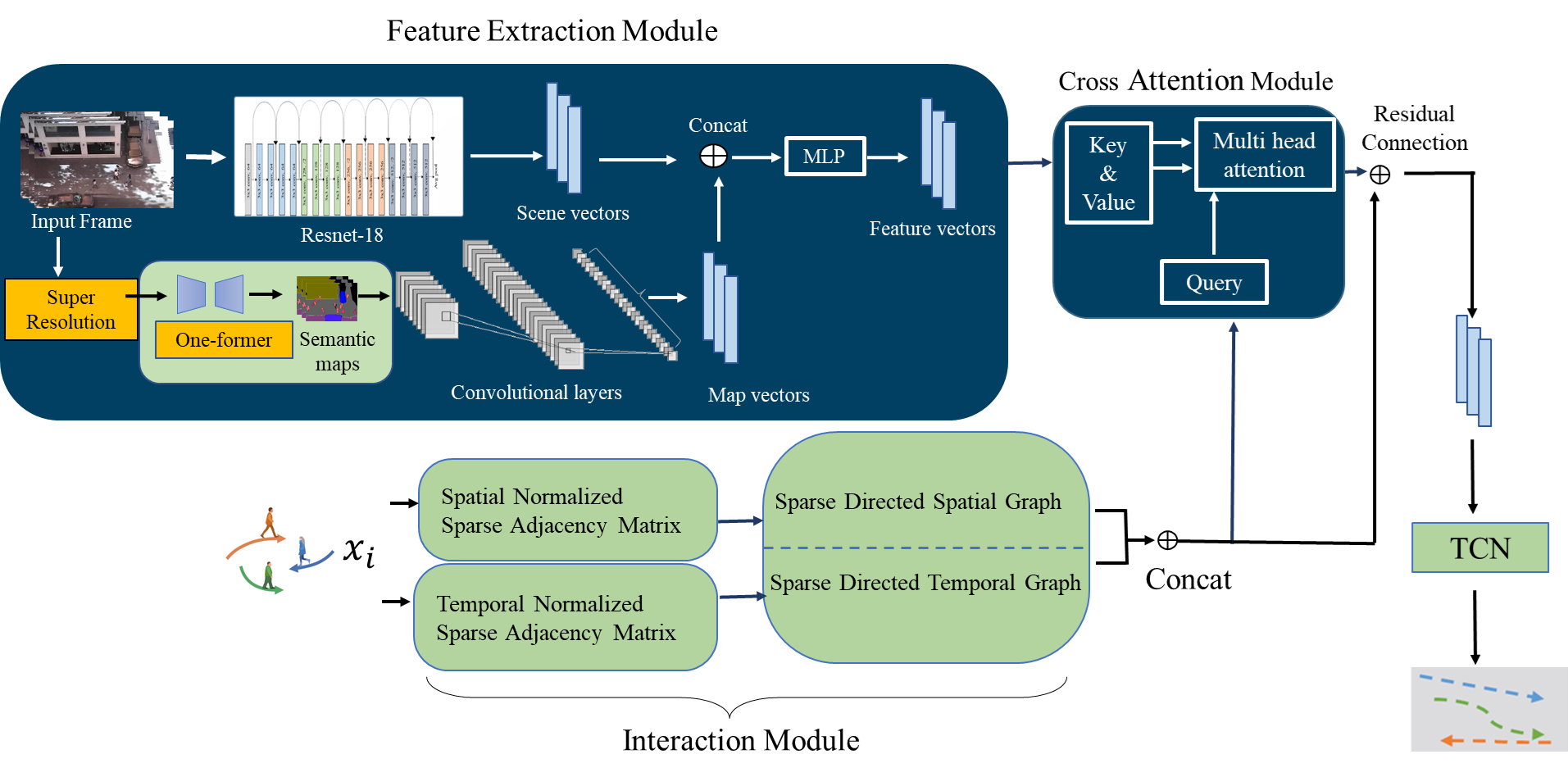}
    \caption{Architecture of our proposed Feature Extraction Module and the pedestrian trajectory prediction pipeline. The feature vectors from both modules are combined using a cross-attention mechanism.}
    \label{fig:1}
\end{figure*}

As illustrated in Figure~\ref{fig:1}, the structure of our Feature Extraction Module and the pedestrian trajectory prediction pipeline are depicted.

\subsection{Trajectory Prediction and Loss Function}

The fused features $H_{\text{fused}}$ are input into the TCN to predict future positions:

\begin{equation}
\hat{\mathbf{Y}} = \text{TCN}(H_{\text{fused}}),
\end{equation}

We employ a composite loss function combining the \textit{Average Displacement Error (ADE)} and the \textit{Final Displacement Error (FDE)} \cite{bae2023eigentrajectory}:

\begin{equation}
\mathcal{L} = \text{ADE} + \text{FDE},
\end{equation}

where:

\begin{equation}
\text{ADE} = \frac{1}{N T} \sum_{i=1}^{N} \sum_{t=1}^{T} \left\| \hat{\mathbf{Y}}_i^t - \mathbf{Y}_i^t \right\|_2,
\end{equation}

\begin{equation}
\text{FDE} = \frac{1}{N} \sum_{i=1}^{N} \left\| \hat{\mathbf{Y}}_i^T - \mathbf{Y}_i^T \right\|_2.
\end{equation}

We train the network end-to-end using stochastic gradient descent, updating all parameters except those of the pre-trained Real-ESRGAN and OneFormer modules, which remain fixed. This allows the model to effectively integrate scene context, social interactions, and temporal dynamics, enhancing prediction accuracy.

\section{Experiments}

We evaluate our proposed method on  pedestrian trajectory datasets and benchmark its performance against established baseline models. Additionally, we assess the quality of our semantic segmentation maps in comparison to those generated by alternative models.

\subsection{Evaluation Datasets}

To validate the effectiveness of our approach, we utilize two prominent public datasets:

\begin{itemize}
    \item \textbf{ETH Dataset}: Comprises two scenes—\textit{ETH} and \textit{HOTEL}.
    \item \textbf{UCY Dataset}: Includes three scenes—\textit{UNIV}, \textit{ZARA1}, and \textit{ZARA2}.
\end{itemize}

Each scene captures real-world pedestrian trajectories characterized by rich social interactions and diverse environmental contexts. Following the standard protocol \cite{alahi2016social}, we adopt a leave-one-out evaluation strategy, training the model on four scenes and testing on the remaining one in a rotating manner. Specifically, we observe pedestrian trajectories for 8 time steps (equivalent to 3.2 seconds) and predict the subsequent 12 time steps (equivalent to 4.8 seconds). Due to the absence of publicly available frame data for the \textit{UNIV} scene (specifically the \texttt{student003} file), we exclude it from frame-based evaluations, conducting our experiments on \textit{ETH}, \textit{HOTEL}, \textit{ZARA1}, and \textit{ZARA2}.

\subsection{Evaluation Metrics}

We employ two standard metrics to quantitatively assess the performance of our trajectory prediction method:

\begin{itemize}
    \item \textbf{Average Displacement Error (ADE)}: Computes the average Euclidean distance between the predicted trajectory points and the ground truth points across all predicted time steps.
    
    \item \textbf{Final Displacement Error (FDE)}: Measures the Euclidean distance between the predicted final position and the ground truth final position at the end of the prediction horizon.
\end{itemize}

\subsection{Results}

As shown in Table~\ref{tab:ade_fde_comparison}, our proposed model, \textbf{ScenePTP}, achieves superior performance across all datasets, outperforming baseline models in both ADE and FDE metrics. Specifically, our model reduces ADE and FDE by significant margins compared to previous state-of-the-art methods.
Notably, in the \textit{HOTEL} scene, our model reduces the ADE to 0.145 meters, improving upon the previous best of 0.31 meters achieved by SGCN. This substantial reduction underscores the effectiveness of incorporating detailed scene information and advanced interaction modeling in predicting pedestrian trajectories.

 The combination of advanced image restoration, state-of-the-art semantic segmentation, and sophisticated interaction modeling enables our model to outperform existing methods.
The substantial improvements in ADE and FDE metrics indicate that our model not only predicts trajectories more accurately on average but also achieves better precision in forecasting the final positions of pedestrians. 

Furthermore, as presented in Table~\ref{tab:ablation_semantic_maps}, the model utilizing only frame features exhibits lower accuracy compared to the model that integrates both frame features and semantic maps. This demonstrates that combining semantic maps with frame features provides the model with a more comprehensive understanding of the pedestrians' surrounding environment, thereby enabling more precise trajectory predictions.

\begin{table*}[ht]
\centering
\caption{ADE / FDE comparison across datasets (in meters). Lower values indicate better performance. The best results are highlighted in \textbf{bold}.}
\label{tab:ade_fde_comparison}
\setlength{\tabcolsep}{4pt} 
\renewcommand{\arraystretch}{1.1} 
\small 
\begin{tabular}{lcc|cc|cc|cc|cc}
\hline
\multirow{2}{*}{\textbf{Model}} & \multicolumn{2}{c|}{\textbf{ETH}} & \multicolumn{2}{c|}{\textbf{HOTEL}} & \multicolumn{2}{c|}{\textbf{ZARA1}} & \multicolumn{2}{c|}{\textbf{ZARA2}} & \multicolumn{2}{c}{\textbf{AVG}} \\ \cline{2-11}
 & \textbf{ADE} & \textbf{FDE} & \textbf{ADE} & \textbf{FDE} & \textbf{ADE} & \textbf{FDE} & \textbf{ADE} & \textbf{FDE} & \textbf{ADE} & \textbf{FDE} \\ \hline
Social-LSTM \cite{alahi2016social} & 1.09 & 2.35 & 0.79 & 1.76 & 0.47 & 1.00 & 0.56 & 1.17 & 0.72 & 1.57 \\ \hline
Social-GAN \cite{gupta2018social} & 0.87 & 1.62 & 0.67 & 1.37 & 0.35 & 0.68 & 0.42 & 0.84 & 0.58 & 1.13 \\ \hline
SoPhie \cite{sadeghian2019sophie} & 0.70 & 1.43 & 0.76 & 1.67 & 0.30 & 0.63 & 0.38 & 0.78 & 0.54 & 1.13 \\ \hline
Social-STGCNN \cite{mohamed2020social} & 0.64 & 1.11 & 0.49 & 0.85 & 0.34 & 0.53 & 0.30 & 0.48 & 0.44 & 0.74 \\ \hline
SGCN \cite{shi2021sgcn} & 0.57 & 1.00 & 0.31 & 0.53 & 0.29 & 0.51 & 0.22 & 0.42 & 0.35 & 0.62 \\ \hline
Semantic Scene Upgrades \cite{syed2023semantic} & 0.75 & 1.62 & 0.60 & 1.28 & 0.30 & 0.79 & 0.34 & 0.91 & 0.50 & 1.15 \\ \hline
\textbf{Our Model (ScenePTP)} & \textbf{0.419} & \textbf{0.593} & \textbf{0.145} & \textbf{0.212} & \textbf{0.253} & \textbf{0.383} & \textbf{0.191} & \textbf{0.298} & \textbf{0.252} & \textbf{0.372} \\ \hline
\end{tabular}
\end{table*}

\subsection{Ablation Study on Semantic Maps}

To evaluate the impact of incorporating semantic maps into our model, we conduct an ablation study comparing the performance with and without semantic maps. In the first configuration (\textbf{w/o Maps}), we use features extracted directly from the raw frames without any semantic segmentation. In the second configuration (\textbf{w/ Maps}), we enhance the frame features by integrating semantic maps generated from the scene.

Table~\ref{tab:ablation_semantic_maps} presents the ADE and FDE for both configurations across the datasets.

\begin{table}[ht]
\centering
\caption{Ablation study on the effect of using semantic maps. ADE/FDE are reported in meters. Lower values indicate better performance.}
\label{tab:ablation_semantic_maps}
\setlength{\tabcolsep}{4pt} 
\renewcommand{\arraystretch}{2} 
\footnotesize 
\begin{tabular}{l|c|c|c|c}
\hline
\textbf{Configuration} & \textbf{ETH} & \textbf{HOTEL} & \textbf{ZARA1} & \textbf{ZARA2} \\ \hline
w/o Maps & 0.444 / 0.650 & 0.151 / 0.230 & 0.277 / 0.409 & 0.209 / 0.322 \\ \hline
w/ Maps & \textbf{0.419 / 0.593} & \textbf{0.145 / 0.212} & \textbf{0.253 / 0.383} & \textbf{0.191 / 0.298} \\ \hline
\end{tabular}
\end{table}

These results validate the effectiveness of incorporating semantic maps into our model. The inclusion of semantic maps consistently improves performance across all datasets, with an average reduction in ADE and FDE. This demonstrates that semantic maps provide valuable environmental context, enabling the model to better understand navigable spaces and obstacles, leading to more accurate trajectory predictions.

\subsection{Qualitative Analysis of Semantic Segmentation}

\begin{figure}[ht]
    \centering
    \includegraphics[width=1\linewidth]{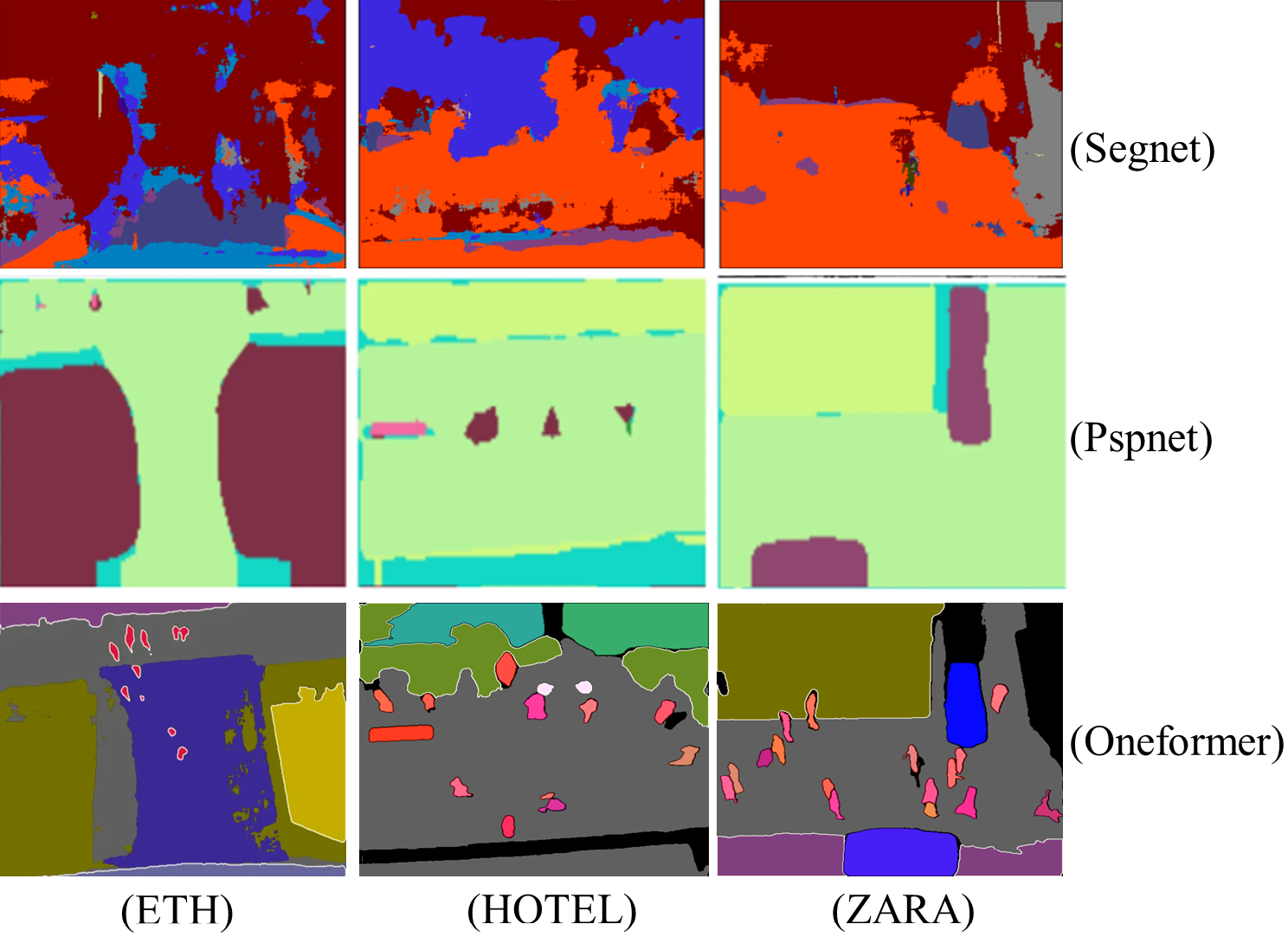}
    \caption{Semantic Segmentation Comparison on ETH, HOTEL, and ZARA Datasets. OneFormer with Image restoration, delivers the most accurate and detailed segmentations, effectively distinguishing complex features better than PSPNet \cite{zhao2017pyramid} and SegNet \cite{badrinarayanan2017segnet}.}
    \label{fig:semantic_segmentation_comparison}
\end{figure}

Figure~\ref{fig:semantic_segmentation_comparison} illustrates an analysis of our semantic maps against those generated by other state-of-the-art semantic segmentation methods, such as PSPNet \cite{zhao2017pyramid} and SegNet \cite{badrinarayanan2017segnet}. The figure highlights the superior accuracy and detail achieved by our approach, particularly in complex urban environments where distinguishing between walkable areas and obstacles is critical for accurate trajectory prediction.

Our semantic segmentation effectively captures intricate details such as narrow pathways and dynamic obstacles, which are often challenging for existing models to accurately delineate. This high-quality semantic information contributes significantly to the improved performance of our trajectory prediction model.

\subsection{Impact of Image Restoration}

\begin{figure}[ht]
    \centering
    \includegraphics[width=1\linewidth]{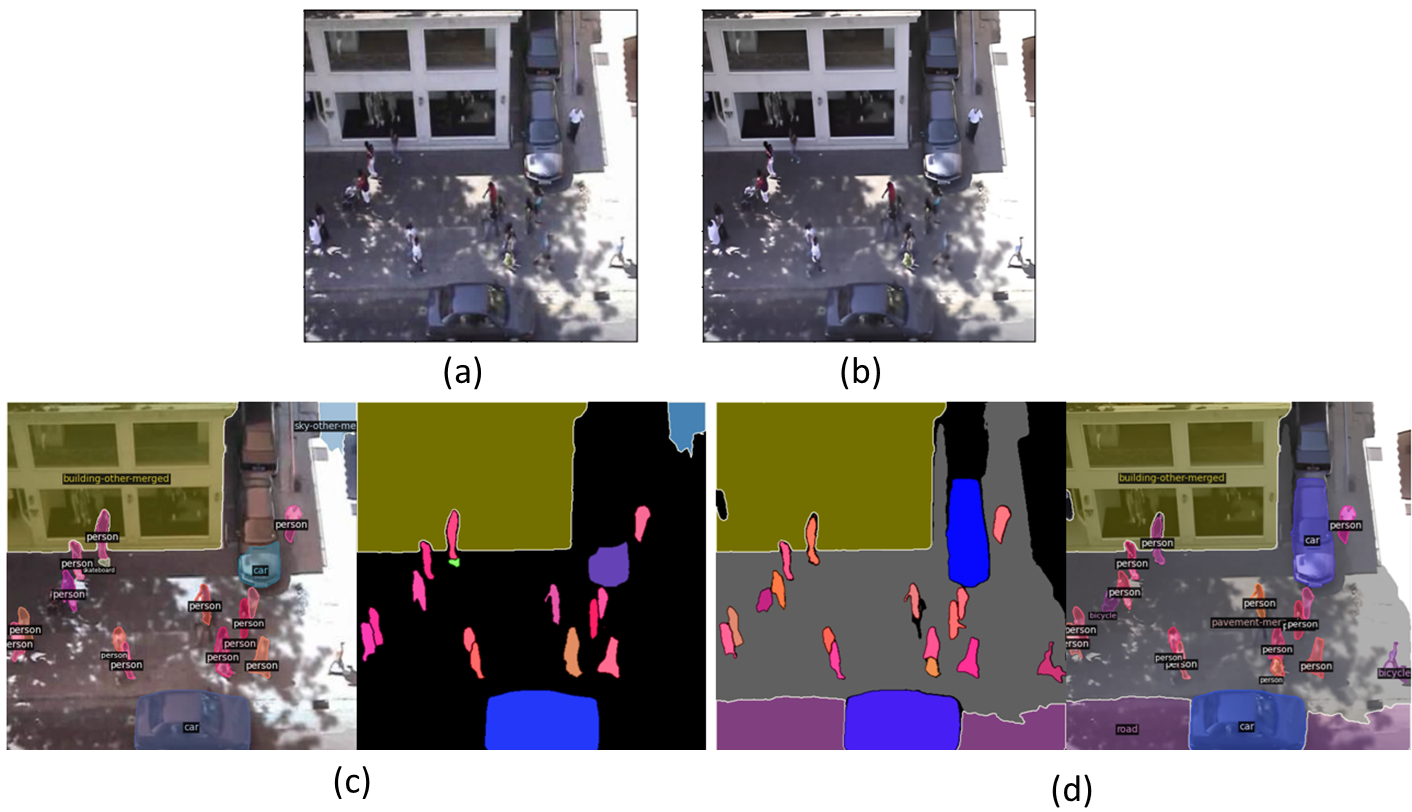}
    \caption{Effect of Image Restoration on Frame Quality and Semantic Segmentation. (a) Original low-resolution frame, (b) Restored high-resolution frame using Real-ESRGAN, (c) Semantic segmentation on the original low-resolution frame, and (d) Improved semantic segmentation on the restored frame. Image restoration enhances visual clarity, leading to more accurate and detailed segmentation results.}
    \label{fig:image_restoration_effect}
\end{figure}

We also assess the effect of image restoration on frame quality and semantic segmentation accuracy. Figure~\ref{fig:image_restoration_effect} shows the comparison between original low-resolution frames and restored high-resolution frames using Real-ESRGAN \cite{wang2021real}. The improved image quality leads to more accurate and detailed semantic segmentation results, as seen in the segmentation maps.

The enhanced segmentation maps provide better environmental context to the model, contributing to the overall improvement in trajectory prediction accuracy.

\subsection{Conclusion}

In this paper, we investigated the impact of incorporating detailed scene information into pedestrian trajectory prediction models. Our approach introduced a novel integration of scene features with graph-based models. By effectively combining environmental context with social interaction modeling, our model significantly improved trajectory prediction accuracy.
Our experimental results validate the effectiveness of our method, highlighting the importance of utilizing environmental context alongside graph representations to achieve more reliable and robust predictions in dynamic environments. The findings demonstrate that integrating scene understanding enhances the model's ability to anticipate pedestrian movements more accurately.

\appendix
\bibliographystyle{ACM-Reference-Format}
\bibliography{sample-base}
\end{document}